\documentclass[10pt,times,twocolumn,final,authoryear]{article}

\usepackage{framed,multirow}

\usepackage{amssymb}
\usepackage{amsmath}
\usepackage{subfig}
\usepackage{graphicx}
\usepackage{latexsym}
\usepackage[colorlinks]{hyperref}

\def\ourmethod{\textit{DIABLO}}

\usepackage{amsmath,amsfonts,bm}

\def\eqref#1{equation~\ref{#1}}

\def\1{\bm{1}}

\def\vd{{\bm{d}}}

\def\vf{{\bm{f}}}

\DeclareMathAlphabet{\mathsfit}{\encodingdefault}{\sfdefault}{m}{sl}
\SetMathAlphabet{\mathsfit}{bold}{\encodingdefault}{\sfdefault}{bx}{n}
\newcommand{\tens}[1]{\bm{\mathcal{#1}}}
\def\tA{{\tens{A}}}

\def\tD{{\tens{D}}}

\def\tF{{\tens{F}}}
\def\tG{{\tens{G}}}
\def\tH{{\tens{H}}}

\def\sR{{\mathbb{R}}}

\DeclareMathOperator*{\argmax}{arg\,max}

\title{DIABLO: Dictionary-based attention block for deep metric learning}
\author{Pierre Jacob$^1$ \quad  David Picard$^2$ \quad  Aymeric Histace$^1$ \quad  Edouard Klein$^3$\\
$^1$ ETIS UMR 8051, Cergy Paris University, ENSEA, CNRS, F-95000, Cergy, France\\
$^2$ LIGM, Ecole des Ponts, Univ Gustave Eiffel, CNRS, Marne-la-Vallée, France\\
$^3$ P\^{o}le Judiciaire de la Gendarmerie Nationale, 5 boulevard de l'Hautil, 95000 Cergy, France}

\begin{document}
\maketitle

\begin{abstract}
    Recent breakthroughs in representation learning of unseen classes and examples have been made in deep metric learning by training at the same time the image representations and a corresponding metric with deep networks.
    Recent contributions mostly address the training part (loss functions, sampling strategies, \emph{etc.}), while a few works focus on improving the discriminative power of the image representation.
    In this paper, we propose \ourmethod, a dictionary-based attention method for image embedding.
    \ourmethod \ produces richer representations by aggregating only visually-related features together while being easier to train than other attention-based methods in deep metric learning.
    This is experimentally confirmed on four deep metric learning datasets (Cub-200-2011, Cars-196, Stanford Online Products, and In-Shop Clothes Retrieval) for which \ourmethod \ shows state-of-the-art performances.
\end{abstract}




\section{Introduction}
    Deep Metric Learning (DML) is an important yet challenging topic in the Computer Vision community, that has a broad-spectrum in terms of applications such as: person or vehicle identification \cite{Zhou_2017_ICCV}, visual product search \cite{Liu_2016_CVPR_INSHOP, Song_2016_CVPR} or multi-modal retrieval \cite{Wehrmann_2018_CVPR, Carvalho_2018_SIGIR}.
    By learning the image representations and an embedding space together, DML methods produce compact representations where visually-related images (\emph{e.g.}, images of the same car model) are close to each other and dissimilar images (\emph{e.g.}, images of two cars from the same brand but from different models) are distant.
    
    Recent contributions mainly address the training part of deep metric learning, proposing loss functions (\emph{e.g.} Angular loss \cite{Wang_2017_ICCV}), sampling strategies (\emph{e.g.}, DAMLRMM \cite{Xu_2019_CVPR}) and ensemble methods (\emph{e.g.}, BIER \cite{Opitz_2017_ICCV}).
    All of these methods are built upon a backbone network such as GoogleNet \cite{Szegedy_2015_CVPR} to extract the local features from which the image representations and the corresponding metric are computed.
    Nowadays, global average pooling is the most used pooling strategy to compute the image representations.
    This is due to interesting properties such as full back-propagation of the gradient  or localization ability without direct supervision \cite{Zhou_2016_CVPR}.
    In almost every method, these representations are computed using the mean of the deep features and used as they are.
    
    However, the average is also known to be a non-robust representation because it is very sensitive to outliers and to sampling problems \cite{Jacob_2019_ICCV}.
    A famous solution is to strengthen this representation in order to compute the average on only visually-related features, using a set of attention maps such as ABE \cite{Kim_2018_ECCV}.
    ABE is based on what we next call \emph{dimension-wise selection with pre-attention} - this method shows very good results with few additional parameters, through both pooling of visually-related features and feature denoising.
    Also, ABE is trained with a divergence loss which ensures that the attention maps are complementary by enforcing two attention maps to be dissimilar, even for visually-related images.
    Due to this optimization criterion and the trade-off parameter, both the training procedure and the parameterization become more complex.
    NetVLAD \cite{Arandjelovic_2016_CVPR} is based on what we next call \emph{feature-wise selection with post-attention} - this method aggregates visually-similar features using a structural constraint based on a dictionary strategy.
    However, the \emph{feature-wise selection} does not contribute to feature denoising.
    
    In this paper, we introduce the method \ourmethod, a DIctionary-based Attention BLOck, that produces robust image representations by taking advantage of both ABE and NetVLAD benefits.
    We evaluate attention strategies named \emph{pre-attention} and \emph{post-attention} in the DML setup, together with two selection strategies named \emph{feature-wise} and \emph{dimension-wise} selection.
    We show in practice that \ourmethod \ consistently improves the state-of-the-art on four DML datasets (Cub-200-2011, Cars-196, Stanford Online Products and In-Shop Clothes Retrieval).
    
    The remaining of the paper is organized as follows: in the next section, we present the related work on deep metric learning and support our proposition.
    In \autoref{sec:met}, we give an overview of the proposed architecture, then we detail the attention strategies and the feature-wise and dimension-wise selection methods.
    In \autoref{sec:abla}, we show ablation studies on the selection and attention strategies, the dictionary size and the comparison to ABE.
    Finally, in \autoref{sec:results} we compare our approach to the state-of-the-art methods on four image retrieval datasets (Cub-200-2011, Cars-196, Stanford Online Products and In-Shop Clothes Retrieval).
    
    \begin{figure*}[t]
        \centering
        \subfloat[Post-attention strategy\label{fig:post_att}]{\includegraphics[width=0.43\linewidth]{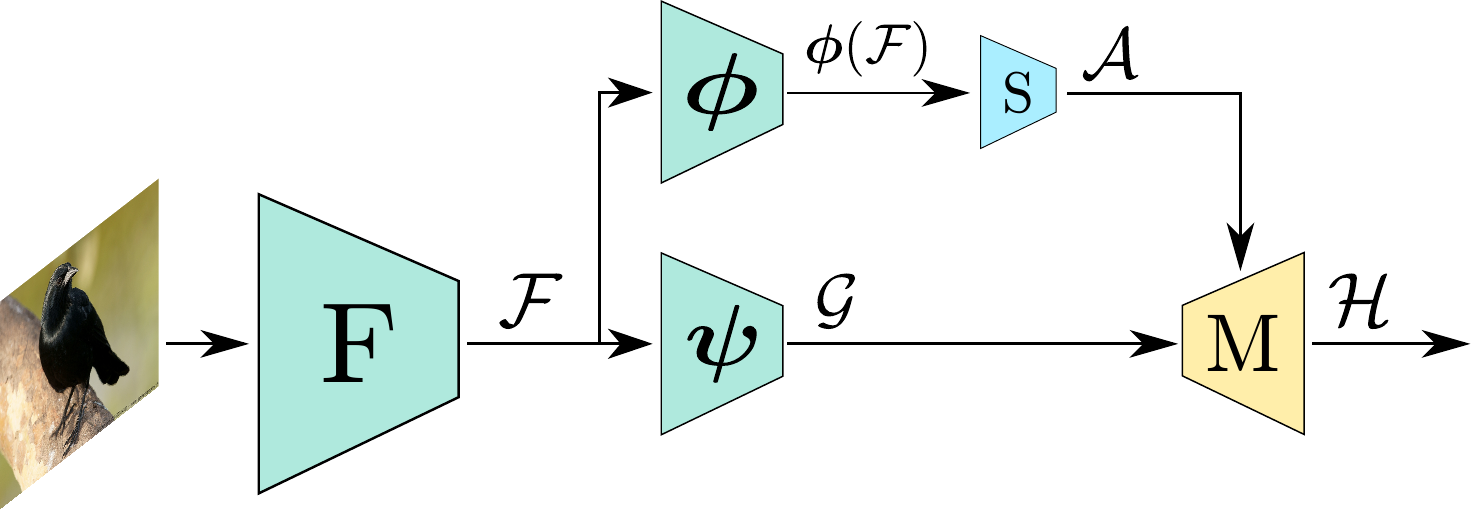}} \hfill
        \subfloat[Pre-attention strategy\label{fig:pre_att}]{\includegraphics[width=0.5\linewidth]{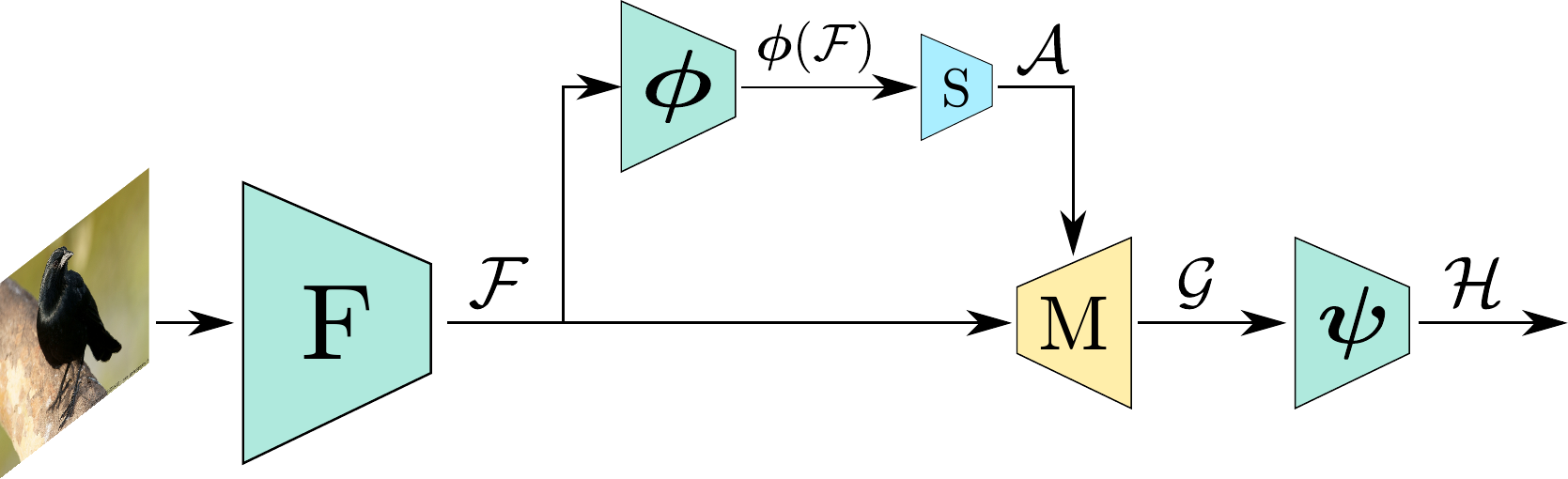}}
        \caption{Illustration of post and pre-attention strategies. In pre-attention, we first compute the attention maps and then we process each of them. In post-attention, the feature map is further processed before computing the attention maps. In pre-attention, we first compute the attention maps and then we process each of them. $F$ is a feature extractor and $\boldsymbol{\phi}$ and $\boldsymbol{\psi}$ are two non-linear transformation of the deep features. The blocks $\mathcal{S}$ correspond to either \autoref{eq:feat_sel} or \autoref{eq:dim_sel}. The blocks $\mathcal{M}$ are illustrated in \autoref{fig:selection_strategies}.}
        \label{fig:diablo_architecture}
    \end{figure*}
        
    \begin{figure}[t]
        \centering
        \subfloat[Feature-wise selection\label{fig:feat_sel}]{\includegraphics[width=0.475\linewidth]{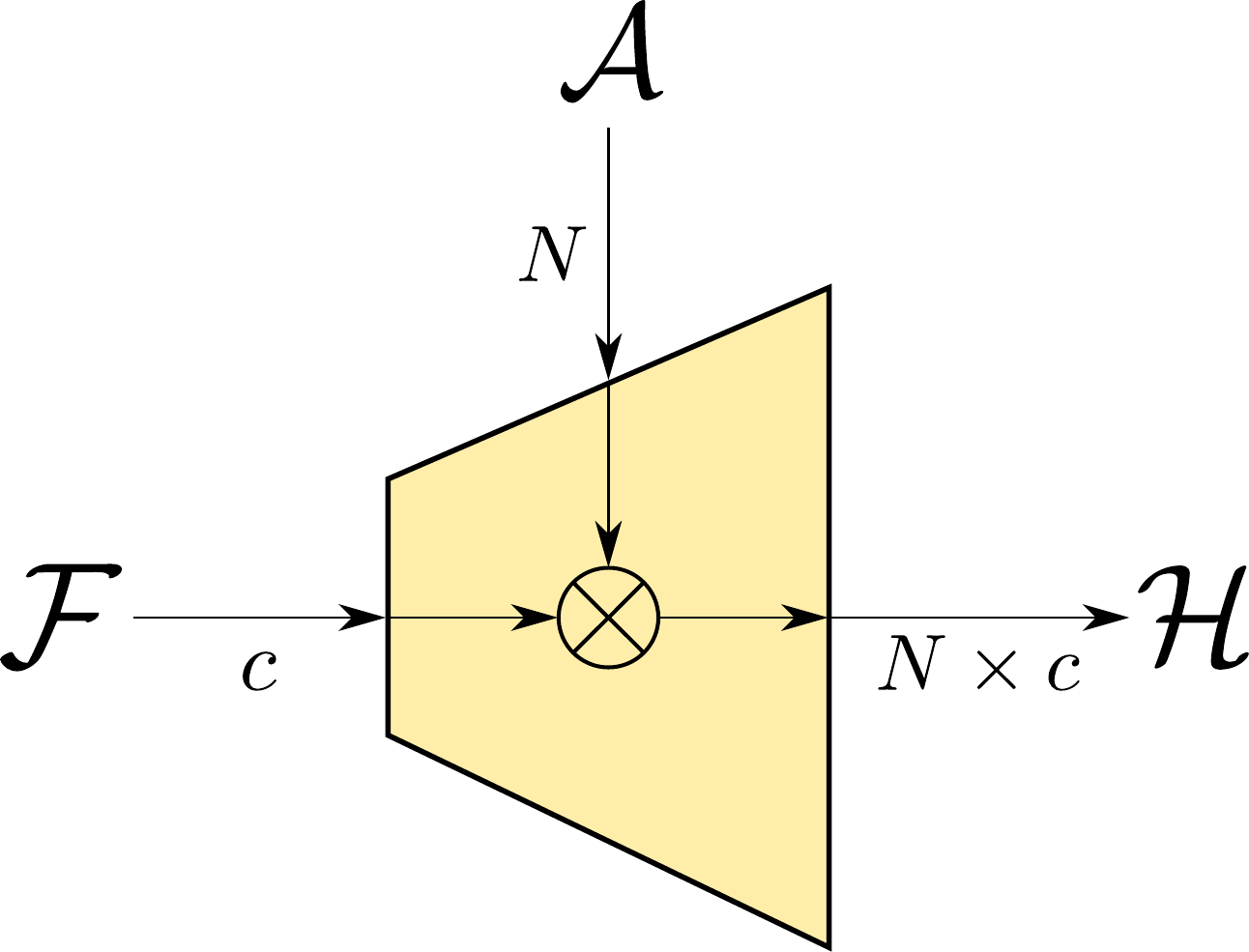}}
        \hfill
        \subfloat[Dimension-wise selection\label{fig:dim_sel}]{\includegraphics[width=0.475\linewidth]{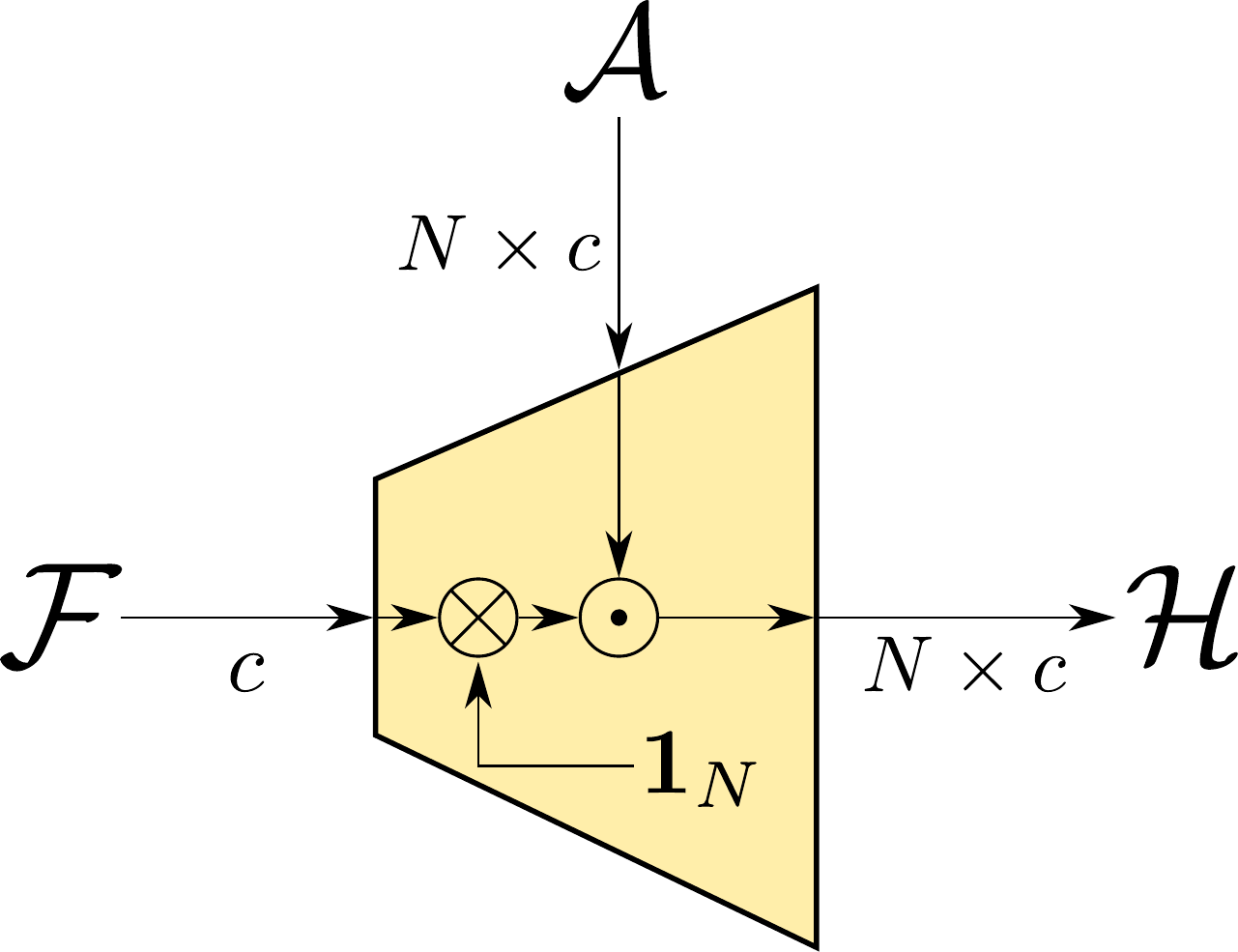}}
        \caption{Illustration of the merging block M for feature-wise and dimension-wise selection strategies. The feature-wise selection is directly implemented using Kronecker product ($\otimes$). The dimension-wise selection first duplicate $N$ times the feature map before computing the Hadamard product ($\odot$) with the attention maps. Reported dimensions are without the spatial dimensions $h \times w$.}
        \label{fig:selection_strategies}
    \end{figure}

\section{Related work}
    In DML, we learn the image representations and an embedding together in such a way that the Euclidean distance corresponds with the semantic content of the images.
    The standard strategy is to extract deep features using a backbone network such as GoogleNet \cite{Szegedy_2015_CVPR} and to learn the target representation with a linear projection on the average of these deep features.
    The whole network is fine-tuned to solve the metric learning task according to three criteria: a similarity-based loss function, a sampling strategy, and an ensemble method.
    
    Standard loss functions rely on pairs or triplets of similar/dissimilar samples.
    Most recent ones extend these formulations by considering larger tuples \cite{Ustinova_2016_NIPS} or by improving the design \cite{Wang_2017_ICCV}.
    The batch construction can either be done by random sampling, mining strategies \cite{Xu_2019_CVPR}, proxy-based approximations \cite{Movshovitz-Attias_2017_ICCV} or generation \cite{Lin_2018_ECCV}.
    Finally, ensemble methods have recently become a popular way of improving the performances of DML architectures \cite{Opitz_2017_ICCV, Opitz_toap_PAMI}.
    
    The latest DML approaches consider using a codebook strategy \cite{Arandjelovic_2016_CVPR} or even attention maps \cite{Kim_2018_ECCV}.
    \cite{Kim_2018_ECCV} propose ABE, an attention-based ensemble method to enforce diversity within the embedding space.
    With this aim, they train a set of attention blocks to select several feature map entries per block.
    These blocks are then trained with a divergence loss function, which ensures that all attention maps are complementary.
    Their proposed design relies on what we call a \emph{dimension-wise selection with pre-attention}.
    they select a set of dimensions from the feature maps using a given attention map.
    Then, these selected dimensions are further processed with Inception blocks.
    We call this selection schema a \emph{pre-attention} strategy - and we argue that one of its properties results in denoising the deep features.
    Thus, this improves the image representations by post-processing features that contain only relevant information.
    However, there are two major drawbacks in using a divergence loss.
    On the one hand, we need to cross-validate the additional trade-off hyper-parameter.
    On the other hand, the training time is increased because the divergence loss plays the opposite role of the similarity-based loss function.
    
    \cite{Arandjelovic_2016_CVPR} propose NetVLAD which takes advantage of a dictionary strategy that avoids the divergence loss.
    In their case, the optimization constraint is replaced by a structural one, that simplifies the training procedure by directly optimizing the task-dependent loss function.
    However, NetVLAD's design relies on a strategy that we call \emph{feature-wise selection with post-attention}.
    Feature-wise selection ensures that only visually-related features are pooled together.
    This selection strategy does not have the denoising property of the \emph{dimension-wise selection}.
    Also, in the case of NetVLAD, the pre-processing is a centering and an intra-projection, that can be improved using non-linear transformation learned by a CNN.
    
    In \ourmethod, we show the benefits of dimension-wise selection versus feature-wise selection with pre-attention or post-attention strategies.
    With this aim, we propose to replace the divergence loss in ABE by a structural constraint, which is based on a dictionary strategy.
    This dictionary is trained without direct supervision as it is the case with NetVLAD.
    We show that it leads to better results than ABE on four DML datasets.

\section{Proposed method}\label{sec:met}
    In this section, we start by giving an overview of \ourmethod.
    Then, we describe the attention maps computation and the two selection strategies named feature-wise and dimension-wise selection.
    
    \subsection{Method overview}\label{sec:met_overview}
        We give an overview of the method illustrated in \autoref{fig:diablo_architecture}.
        We start by extracting a deep feature map $\tF \in \sR^{h \times w \times c}$ using the local feature extractor F where $h$ and $w$ are the height and width of the feature map and $c$ is the deep feature dimension.
        We further process the extracted feature map using a non-linear function $\boldsymbol{\phi}$ implemented by a convolutional neural network (CNN).
        In order to compute the $N$ attention maps $\tA$ where $N$ is the dictionary size, we pass the feature map into the selection block S, that is either the feature-wise selection (\autoref{sec:met_feat}) or the dimension-wise selection (\autoref{sec:met_dim}).
        
        In the post-attention setup (\autoref{fig:post_att}), the feature map $\mathcal{F}$ is further processed with a non-linear function $\boldsymbol{\psi}$ (implemented by a CNN) and transformed into a feature map $\mathcal{G}$.
        It is then combined with the attention maps $\tA$ using the block M to produce the $N$ new feature maps.
        In the pre-attention setup (\autoref{fig:pre_att}), the feature map $\mathcal{F}$ is directly combined with the attention maps $\tA$ using the block M.
        In this case, we further process these $N$ feature maps using a non-linear function $\boldsymbol{\psi}$.
        The different blocks M used to combine the attention maps and the original feature map are illustrated in \autoref{fig:selection_strategies}.
        
        These $N$ feature maps are pooled using a global average pooling with adding an embedding layer for each branch.
        So, the output representation is obtained by concatenating these $N$ branches.
        
    \subsection{Attention strategies}\label{sec:met_mdl}
        This section focuses on computing a set of attention maps $\tA$ using one of the selection blocks S and its corresponding dictionary $\tD$ as well as the feature map $\tF \in \sR^{h \times w \times c}$.
        For this purpose, we process the feature map $\tF$ with a non-linear function $\boldsymbol{\phi}: \sR^c \xrightarrow{} \sR^m$ implemented by a CNN.
        Then, the set of attention maps are computed by a selection block S such that $\tA = \text{S}(\boldsymbol{\phi}(\tF) ; \tD)$.
        $\tA$ is used with one of the following attention strategies.
        
        \subsubsection{Post-attention}
            In the post-attention strategy illustrated in \autoref{fig:post_att}, we process the feature map $\tF$ into an intermediate feature map $\tG$:
            \begin{align}
                \tG = \boldsymbol{\psi}\left(\tF\right)
            \end{align}
            Then, we combine this feature map $\tG$ with the attention maps $\tA$ using the merging block M:
            \begin{align}
                \tH = M(\tG ; \tA)
            \end{align}
            with $\tH$ representing the set of feature maps obtained by the post-attention strategy.
            From these, we perform a spatial pooling of the local features and we add an embedding layer to generate the corresponding image representation.
            
            The main idea of post-attention consists of aggregating only the related features, unlike global pooling that aggregates unrelated features.
            For example, features that describe the background are aggregated using a different attention block than those which represent the desired object.
            One can note that this approach is strongly related to NetVLAD for which the function $\boldsymbol{\psi}$ is a centering.
            However, it differs from NetVLAD in two ways: First, because it learns a non-linear clustering of the local features using the function $\boldsymbol{\phi}$; second because it learns a non-linear pre-processing of the deep features using the function $\boldsymbol{\psi}$.
            
        \subsubsection{Pre-attention}
            In the pre-attention strategy illustrated in \autoref{fig:pre_att}, we perform the operation in reverse order.
            In order to do so, we combine the feature map $\tF$ with the set of attention maps $\tA$ using the merging block M to produce the set of $N$ (one per attention block) feature maps $\tG$:
            \begin{align}
                \tG = M(\tF ; \tA)
            \end{align}
            Then, we process these $N$ feature maps using a non-linear function $\boldsymbol{\psi}$ to generate the feature maps $\tH$:
            \begin{align}
                \tH = \boldsymbol{\psi}\left(\tG\right)
            \end{align}
            From these $N$ feature maps, we pool the local features and we add an embedding layer to generate the corresponding image representation.
            
            The pre-attention strategy is similar to a refinement approach.
            Indeed, the attention selects dimensions or features, and a refinement function $\boldsymbol{\psi}$ improves these extracted features before that they are aggregated.
            Hence, the attention maps role is to select only the relevant information for a given attention block.
            The function $\boldsymbol{\psi}$ is trained to refine this information so that it improves generalization.
            
    \subsection{Feature-wise selection}\label{sec:met_feat}
        In this section, we give details about the computation of the selection block S and the merging block M for the feature-wise selection illustrated in \autoref{fig:feat_sel}.
        We start from a feature map $\tF \in \sR^{h \times w \times c}$ from which we compute a set of attention maps $\tA = \{ \tA^{(n)} \in \sR^{h \times w \times c} \}_n$ using a dictionary $\mathcal{D}$ such that $\tA = \text{S}(\tF ; \mathcal{D})$.
        Before computing the feature-wise selection, the feature map $\tF$ is processed by a non-linear function $\boldsymbol{\phi}: \sR^c \xrightarrow{} \sR^m$ implemented by a CNN.
        We denote $\vf_{i,j} \in \sR^c$ a feature from $\tF$ at spatial location $(i,j)$ and $\boldsymbol{\phi}(\vf_{i,j}) \in \sR^m$ its transformation in $\boldsymbol{\phi}(\tF)$ that is in the same spatial location.
        
        \subsubsection{Selection block}\label{sec:met_feat_sel}
            In the feature-wise selection, the objective is to assign each feature $\vf_{i,j}$ from the feature map $\tF$ to one of the dictionary entries $\tD = \{ \vd^{(n)} \in \sR^m \}_n$.
            To do so, we consider the cosine similarity between the transformed feature $\boldsymbol{\phi}(\vf_{i,j})$ and the dictionary entry:
            \begin{align}
                s(\boldsymbol{\phi}(\vf_{i,j}), \vd^{(n)}) = \frac{\left< \boldsymbol{\phi}(\vf_{i,j}) ~;~ \vd^{(n)} \right>}{\| \boldsymbol{\phi}(\vf_{i,j}) \|_2 \| \vd^{(n)} \|_2}
            \end{align}
            Using the cosine similarity has a main advantage when compared to the Euclidean distance: during the training, it is more stable by means of the $\ell_2$-normalization.
            
            From this similarity measure, we compute $\tA^{(n)}_{i,j,k}$, the weight for the $n$-th feature-wise attention map and $k$-th dimension of the feature in spatial location $(i,j)$:
            \begin{align}
                \tA^{(n)}_{i,j,k} = \left\{
                                        \begin{array}{ll}
                                            1 & \text{if} \ \vd^{(n)} = \underset{\vd^{(l)} \in \ \tD}{\argmax} \ s(\boldsymbol{\phi}(\vf_{i,j}), \vd^{(l)})\\
                                            0 & \text{else.}
                                        \end{array}
                                    \right.
            \end{align}
            However, this one-hot encoder is not differentiable due to the $\argmax$ operator.
            We relax the constraint using the soft-max operator to train in an end-to-end way the deep network:
            \begin{align}\label{eq:feat_sel}
                \tA^{(n)}_{i,j, k} = \frac{{\rm e}^{\alpha \ s(\boldsymbol{\phi}(\vf_{i,j}), \vd^{(n)})}}{\sum_l {\rm e}^{\alpha \ s(\boldsymbol{\phi}(\vf_{i,j}), \vd^{(l)})}}
            \end{align}
            such that, $\alpha$ is a hyper-parameter to control the hardness of the assignment.
            For this formulation, we rely on a given feature $\vf_{i,j}$ that is assigned to a dictionary entry $\vd^{(n)}$ according to the similarity between $\boldsymbol{\phi}(\vf_{i,j})$ and $\vd^{(n)}$.
            We show in \autoref{sec:results} that the feature-wise selection increases the performance of the attention-based models when compared to the baseline model (without attention map).
            Then, we detail the block M to merge the feature-wise selection based attention map with the raw features.
            
        \subsubsection{Merging block}\label{sec:met_feat_mrg}
            The combination of the attention maps $\tA$ and the feature map $\tF$ is illustrated in \autoref{fig:feat_sel}.
            For the $k$-th dimension of the $n$-th feature map in spatial location $(i,j)$, the corresponding entry in $\tH$, $\tH^{(n)}_{i,j,k}$ is computed using the following equation:
            \begin{align}\label{eq:mrg_feat}
                \tH^{(n)}_{i,j,k} = \tA^{(n)}_{i,j,k} \ \tF_{i,j,k}
            \end{align}
            It must be considered that $\tA^{(n)}_{i,j,k}$ has the same value independently from the value of $k$.
            Thus, \autoref{eq:mrg_feat} is easily implemented using the Kronecker product ($\otimes$) as illustrated in \autoref{fig:feat_sel}:
            \begin{align}
                \tH_{i,j} = \tA_{i,j} \otimes \tF_{i,j}
            \end{align}
            where $\tA_{i,j}$ is composed by the $N$-th assignment weights and $\vf_{i,j} \in \sR^c$ is the feature for all spatial locations $(i,j)$.
            
    \subsection{Dimension-wise selection}\label{sec:met_dim}
        In this section, we extend the feature-wise selection from the \autoref{sec:met_feat} to the dimension-wise selection.
        Similarly, we give details about the attention maps computation through the selection block S before that we explain the merging strategy with the block M.
    
        \subsubsection{Selection block}\label{sec:met_dim_sel}
            The selection block is composed of a set of directions per dictionary entry $\tD = \{ \vd^{(n)}_k \in \sR^{m} \}_{n,k}$ of size $N \times c$, on the contrary to the feature-wise selection that has a dictionary $\tD = \{ \vd^{(n)} \in \sR^m \}_n$ of size $N$.
            Thus, for a given feature $\vf_{i,j} \in \tF$ in spatial location $(i, j)$, the cosine similarity is computed between the transformed feature $\boldsymbol{\phi}(\vf_{i,j})$ and the $k$-th direction of the $n$-th dictionary entry $\vd^{(n)}_k \in \tD$:
            \begin{align}
                s(\boldsymbol{\phi}(\vf_{i,j}), \vd^{(n)}_k) = \frac{\left< \boldsymbol{\phi}(\vf_{i,j}) ~;~ \vd^{(n)}_k \right>}{\| \boldsymbol{\phi}(\vf_{i,j}) \|_2 \| \vd^{(n)}_k \|_2}
            \end{align}
            Then, one entry $\tA^{(n)}_{i,j,k}$ from the attention map $\tA$ is computed using the following equation:
            \begin{align}\label{eq:dim_sel}
                \tA^{(n)}_{i,j,k} = \frac{{\rm e}^{\alpha \ s(\boldsymbol{\phi}(\vf_{i,j}), \vd^{(n)}_k)}}{\sum_l {\rm e}^{\alpha \ s(\boldsymbol{\phi}(\vf_{i,j}), \vd^{(l)}_k)}}
            \end{align}
            The attention map has an attention weight for each dimension of the input feature and for each of the $N$ attention blocks.
            \autoref{sec:results} highlights that the dimension-wise selection produces stronger image representations than the feature-wise selection, showing much higher performances.
        
        \subsubsection{Merging block}\label{sec:met_dim_mrg}
            The merging block M is used in the case of dimension-wise selection as illustrated in \autoref{fig:dim_sel}.
            The entry $\tH^{(n)}_{i,j,k}$ in $\tH$ is computed for the $k$-th dimension of the $n$-th feature map in the spatial location $(i,j)$ using the following equation:
            \begin{align}\label{eq:mrg_dim}
                \tH^{(n)}_{i,j,k} = \tA^{(n)}_{i,j,k} \ \tF_{i,j,k}
            \end{align}
            Note that $\tA^{(n)}_{i,j,k}$ depends on the value of $k$.
            The computation can easily be re-written using the Kronecker product and the Hadamard product ($\odot$) as illustrated in \autoref{fig:dim_sel}.
            Using the Kronecker product, we duplicate $N$ times the local feature $\vf_{i,j}$
            Then, $\tH$ is computed using the element-wise product between $\tA$ and the duplicated feature map:
            \begin{align}
                \tH = \left( \tF \otimes \mathbf{1}_N \right) \odot \tA
            \end{align}

    \subsection{Implementation details} \label{sec:met_imp}
        For a fair comparison with other methods, we use a pre-trained GoogleNet \cite{Szegedy_2015_CVPR} on ImageNet.
        The embedding size is fixed to 512 for all models.
        As it is done in common practice, we set the triplet margin $\alpha = 0.1$, the contrastive and the binomial margin $\beta = 0.5$ and the negative sample weight $C=25$ in the binomial deviance.
        We follow the same training procedure as state-of-the-art methods \cite{Opitz_2017_ICCV, Kim_2018_ECCV}: training hyper-parameters (learning rate, batch size, \emph{etc.}) are empirically chosen based on the training loss after a few epochs.
        Model hyper-parameters (number of layers, \emph{etc.}) are set to comparable values as the ones of models reported in ABE \cite{Kim_2018_ECCV}.
        We use the following data augmentation on the images: multi-resolution where the size is uniformly sampled in $[0.8, 1.8]$ times the crop size as well as random $256 \times 256$ crop and horizontal flip during the training.
        During testing, we re-scale the images to $256 \times 256$.
        We use Adam \cite{Diederik_2015_ICLR} with the default parameters and a learning rate of $10^{-5}$.
        For all models, the function $\boldsymbol{\psi}$ is shared across the attention maps to reduce the large number of parameters induced.
        In practice, it still leads to strong experimental results and avoids over-fitting on small datasets, for instance: Cub-200-2011 and Cars-196.

\section{Ablation studies}\label{sec:abla}
    In this section, we compare our approach to the baseline in terms of model complexity and computation.
    Then, we present the evaluation protocol of our ablation studies.
    Ablation studies are performed to evaluate the benefits of pre and post-attention, the assignment approaches (feature and dimension-wise), the dictionary strategy and the dictionary size.
        
        \setlength{\tabcolsep}{5pt}
    \begin{table*}[t!]
        \footnotesize
        \begin{center}
            \caption{Comparison with the state-of-the-art on Cub-200-2011 and Cars-196 datasets using GoogleNet as feature extractor. Results are in percents. State-of-the-art results are in bold and results which improve the baseline are underlined.}
            \label{tab:CUB-CARS}
            \begin{tabular}{|c|cccccc|cccccc|}\hline
                 & \multicolumn{6}{c|}{Cub-200-2011} & \multicolumn{6}{c|}{Cars-196} \\\hline
                R@ & 1 & 2 & 4 & 8 & 16 & 32 & 1 & 2 & 4 & 8 & 16 & 32 \\\hline
                RML \cite{Roy_2019_ICCV} & 52.3 & 64.5 & 75.3 & 84.0 & - & - & 73.2 & 82.2 & 88.6 & 92.2 & - & - \\
                Angular loss \cite{Wang_2017_ICCV} & 54.7 & 66.3 & 76.0 & 83.9 & - & - & 71.4 & 81.4 & 87.5 & 92.1 & - & - \\
                DAML \cite{Duan_2018_CVPR} & 52.7 & 65.4 & 75.5 & 84.3 & - & - & 75.1 & 83.8 & 89.7 & 93.5 & - & - \\
                HDML \cite{Zheng_2019_CVPR} & 53.7 & 65.7 & 76.7 & 85.7 & - & - & 79.1 & 87.1 & 92.1 & 95.5 & - & - \\
                DAMLRMM \cite{Xu_2019_CVPR} & 55.1 & 66.5 & 76.8 & 85.3 & - & - & 73.5 & 82.6 & 89.1 & 93.5 & - & - \\
                HDC \cite{Yuan_2017_ICCV} & 53.6 & 65.7 & 77.0 & 85.6 & 91.5 & 95.5 & 73.7 & 83.2 & 89.5 & 93.8 & 96.7 & 98.4 \\
                BIER \cite{Opitz_2017_ICCV} & 55.3 & 67.2 & 76.9 & 85.1 & 91.7 & 95.5 & 78.0 & 85.8 & 91.1 & 95.1 & 97.3 & 98.7 \\
                DVML \cite{Lin_2018_ECCV} & 52.7 & 65.1 & 75.5 & 84.3 & - & - & 82.0 & 88.4 & 93.3 & 96.3 & - & - \\
                HTG \cite{Zhao_2018_ECCV} & 59.5 & 71.8 & 81.3 & 88.2 & - & - & 76.5 & 84.7 & 90.4 & 94.0 &- & - \\
                HTL \cite{Ge_2018_ECCV} & 57.1 & 68.8 & 78.7 & 86.5 & 92.5 & 95.5 & 81.4 & 88.0 & 92.7 & 95.7 & 97.4 & 99.0 \\
                A-BIER \cite{Opitz_toap_PAMI} & 57.5 & 68.7 & 78.3 & 86.2 & 91.9 & 95.5 & 82.0 & 89.0 & 93.2 & 96.1 & 97.8 & 98.7 \\
                JCF \cite{Jacob_2019_ICIP} & 60.1 & 72.1 & 81.7 & 88.3 & - & - & 82.6 & 89.2 & 93.5 & 96.0 & - & - \\
                HORDE \cite{Jacob_2019_ICCV} & 59.4 & 71.0 & 81.0 & 88.0 & 93.1 & 96.5 & 83.2 & 89.6 & 93.6 & 96.3 & 98.0 & 98.8 \\
                ABE \cite{Kim_2018_ECCV} & 60.6 & 71.5 & 79.8 & 87.4 & - & - & 85.2 & 90.5 & 94.0 & 96.1 & - & - \\
                \hline
                Contrastive (Ours) & 58.7 & 69.7 & 79.4 & 87.0 & 92.6 & 96.1 & 78.5 & 85.9 & 90.9 & 94.4 & 96.7 & 98.1 \\
                Contrastive + \ourmethod & \underline{62.3} & \underline{73.6} & \underline{82.6} & \underline{89.2} & \underline{94.0} & \underline{96.9} & \underline{84.8} & \underline{90.5} & \underline{94.3} & \underline{96.6} & \underline{98.1} & \underline{98.9} \\
                Triplet (Ours) & 55.9 & 67.0 & 77.7 & 86.1 & 92.0 & 95.5 & 73.1 & 81.7 & 87.9 & 92.9 & 95.9 & 97.6 \\
                Triplet + \ourmethod &\underline{59.6} & \underline{70.6} & \underline{80.3} & \underline{87.7} & \underline{92.7} & \underline{96.2} & \underline{75.0} & \underline{83.4} & \underline{89.4} & \underline{93.5} & \underline{96.4} & \underline{98.1} \\
                Binomial (Ours) & 59.6 & 70.8 & 81.0 & 88.1 & 93.1 & 96.2 & 78.8 & 86.2 & 91.5 & 94.8 & 97.0 & 98.4 \\
                Binomial + \ourmethod \ $N=8$ & \underline{62.8} & \underline{73.9} & \underline{82.4} & \underline{89.3} & \underline{94.0} & \underline{96.7} & \underline{85.0} & \underline{90.8} & \underline{94.0} & \underline{96.4} & \underline{98.0} & \underline{98.9} \\
                Binomial + \ourmethod \ $N=16$ & \textbf{63.9} & \textbf{74.3} & \textbf{82.4} & \textbf{88.8} & \textbf{94.0} & \textbf{96.8} & \textbf{85.4} & \textbf{91.3} & \textbf{95.0} & \textbf{97.2} & \textbf{98.5} & \textbf{99.1} \\
                \hline
            \end{tabular}
        \end{center}
    \end{table*}

    \setlength{\tabcolsep}{7pt}
    \begin{table*}[t!]
        \footnotesize
        \begin{center}
            \caption{Comparison with the state-of-the-art on Stanford Online Products and In-Shop Clothes Retrieval using GoogleNet as feature extractor. Results in percents. State-of-the-art results are in bold and results which improve the baseline are underlined.}
            \begin{tabular}{|c|cccc|cccccc|}\hline
                 & \multicolumn{4}{c|}{Stanford Online Products} & \multicolumn{6}{c|}{In-Shop Clothes Retrieval} \\\hline
                R@ & 1 & 10 & 100 & 1000 & 1 & 10 & 20 & 30 & 40 & 50 \\\hline
                Angular loss \cite{Wang_2017_ICCV} & 70.9 & 85.0 & 93.5 & 98.0 & - & - & - & - & - & - \\
                DAML \cite{Duan_2018_CVPR} & 68.4 & 83.5 & 92.3 & -  & - & - & - & - & - & - \\
                HDML \cite{Zheng_2019_CVPR} & 68.7 & 83.2 & 92.4 & - & - & - & - & - & - & - \\
                RML \cite{Roy_2019_ICCV} & 69.2 & 83.1 & 92.7 & - & - & - & - & - & - & - \\
                DAMLRMM \cite{Xu_2019_CVPR} & 69.7 & 85.2 & 93.2 & - & - & - & - & - & - & - \\
                HDC \cite{Yuan_2017_ICCV} & 69.5 & 84.4 & 92.8 & 97.7 & 62.1 & 84.9 & 89.0 & 91.2 & 92.3 & 93.1 \\
                BIER \cite{Opitz_2017_ICCV} & 72.7 & 86.5 & 94.0 & 98.0 & 76.9 & 92.8 & 95.2 & 96.2 & 96.7 & 97.1 \\
                DVML \cite{Lin_2018_ECCV} & 70.2 & 85.2 & 93.8 & - & - & - & - & - & - & - \\
                HTG \cite{Zhao_2018_ECCV} & - & - & - & - & 80.3 & 93.9 & 95.8 & 96.6 & 97.1 & - \\
                HORDE \cite{Jacob_2019_ICCV} & 72.6 & 85.9 & 93.7 & 97.9 & 84.4 & 95.4 & 96.8 & 97.4 & 97.8 & 98.1 \\
                HTL \cite{Ge_2018_ECCV} & 74.8 & 88.3 & 94.8 & 98.4 & 80.9 & 94.3 & 95.8 & 97.2 & 97.4 & 97.8\\
                A-BIER \cite{Opitz_toap_PAMI} & 74.2 & 86.9 & 94.0 & 97.8 & 83.1 & 95.1 & 96.9 & 97.5 & 97.8 & 98.0 \\
                ABE \cite{Kim_2018_ECCV} & 76.3 & 88.4 & 94.8 & 98.2 & 87.3 & 96.7 & 97.9 & 98.2 & 98.5 & 98.7 \\
                JCF \cite{Jacob_2019_ICIP} & 77.4 & 89.9 & 95.8 & 98.6 & - & - & - & - & - & - \\
                \hline
                Contrastive (Ours) & 75.0 & 87.9 & 94.5 & 98.1 & 89.0 & 97.2 & 98.0 & 98.4 & 98.6 & 98.7 \\
                Contrastive + \ourmethod \ $N=8$ & \textbf{77.8} & \textbf{89.5} & \textbf{95.3} & \textbf{98.4} & \textbf{91.3} & \textbf{98.1} & \textbf{98.7} & \textbf{99.0} & \textbf{99.1} & \textbf{99.1} \\
                Triplet (Ours) & 70.6 & 85.7 & 94.0 & 98.2 & 85.6 & 96.5 & 97.7 & 98.2 & 98.4 & 98.6 \\
                Triplet + \ourmethod \ $N=8$ & \underline{73.5} & \underline{87.8} & \underline{95.0} & \underline{98.5} & \underline{87.4} & \underline{97.2} & \underline{98.1} & \underline{98.6} & \underline{98.8} & \underline{98.9} \\
                \hline
            \end{tabular}
            \label{tab:SOP-INSHOP}
        \end{center}
    \end{table*}
    \subsection{Model complexity and computation cost}
        In this section, we give the architecture choices to compute the functions $\boldsymbol{\phi}$ and $\boldsymbol{\psi}$ for the post-attention and for the pre-attention architectures.
        Then, we analyze the induced computations and the complexity introduced by the dictionary approach.
        
        In post-attention, the GoogleNet backbone extracts the feature map including and up to the max-pooling between the fourth and the fifth scales.
        To compute $\boldsymbol{\phi}$, we add the two inception blocks named '5a' and '5b' upon these features.
        The function $\boldsymbol{\psi}$ is also composed by two independent inception blocks '5a' and '5b'.
        The attention maps and the weighted features are computed using one of the two proposed strategies.
        Then, each branch is pooled using a global average pooling as well as an embedding layer of size $512/N$ and a $\ell_2$-norm are added.
        The output representation is the concatenation of these $N$ branches which leads to a 512d representation.
        
        In pre-attention, we use the GoogleNet features including and up to the max-pooling between the third and the fourth scales.
        The non-linear function $\boldsymbol{\phi}$ is composed of the five inception blocks from the fourth scale of GoogleNet pre-trained on ImageNet.
        The refinement function $\boldsymbol{\psi}$ is composed by the fourth and fifth scales of GoogleNet, they are shared for each map but they are independent from $\boldsymbol{\phi}$.
        The attention maps and the weighted feature maps are computed using either dimension-wise or feature-wise selection.
        As it the case with pre-attention, each branch is pooled using a global average pooling, an embedding layer of size $512/N$ and a $\ell_2$-norm are added before the concatenation of all branches in order to produce the full 512d representation.
        
        These choices directly follow ABE \cite{Kim_2018_ECCV} and we refer the reader to the related paper for more ablations on the architecture, including the multi-head approach, the attention module and the sharing of parameters across the attention modules.
        
        All additional parameters are included in the computation of the function $\boldsymbol{\phi}$.
        By using the five Inception blocks from the fourth scale of GoogleNet, this leads to 3.5M additional parameters.
        Note that these parameters are shared across the dictionary entries and this drastically reduces the number of parameters.
        Also, note that the function $\boldsymbol{\psi}$ is already included in the number of parameters of GoogleNet.
        
        In terms of computation, the most important additional computations come from the function $\boldsymbol{\psi}$ which is computed on each attention map.
        The computation of the fourth and fifth Inception scales are estimated to 0.7 Gflop (see \cite{Kim_2018_ECCV}, Table 1).
        In comparison, the whole GoogleNet requires 1.6 Gflop to produce the image embedding.
        Note that in the case of the pre-attention, all parameters of $\boldsymbol{\psi}$ are shared across the attention maps, which leads to fewer additional parameters.
        Overall, these choices lead to higher experimental results for both ABE and \ourmethod \ compared to the baseline.
    
    \subsection{Model selection protocol}
        In this section, we detail the evaluation protocol for all ablation studies on the Cub-200-2011 dataset that are performed to select the best model.
        We perform 10 random train-val splits on the training set of Cub-200-2011 for deep metric learning: We randomly choose 50 classes for the training set and we keep the rest of the classes for the validation set.
        Then, we train each model on the training set and we select the model that gives the best performances on the validation set.
        We then compute Recall@K on the testing set of Cub-200-2011 for each train-val split.
        All reported results in \autoref{tab:abla_dictionary} and \autoref{tab:abla_attention} are the average and the standard-deviation of Recall@K on the testing set for the ten runs.
        
    \subsection{Feature selection and attention strategy}\label{sec:abla_featatt}
        First, we evaluate the benefit of the pre and post-attention strategies with respect to the assignment strategy.
        To that end, we fix the dictionary size to 8 and we use the training procedure from \autoref{sec:met_imp}.
        Results are reported in \autoref{tab:abla_attention} for the dataset Cub-200-2011 with binomial loss.
        We remark that all strategies with the exception of the feature-wise selection with pre-attention improve over the baseline and this confirms the benefit of attention maps.
        This experiment also shows that feature-wise attention and dimension-wise attention impact differently the model.
        
        In Feature-wise attention, the post-attention leads to stronger representations with $+1.7\%$ on Recall@1 over the pre-attention.
        We argue that selecting features make better sense with post-attention than pre-attention: only related features are aggregated together with post-attention whereas in pre-attention, the refinement part mostly processes sparse feature maps.
        In the case of dimension-wise attention, both approach provide stronger results with $+3.8\%$ and $+4.3\%$ over the best feature-wise strategy, even though it is still better to use the pre-attention approach.
        We argue that pre-attention is better with dimension-wise selection because the refinement part processes denoised features.
        Indeed, certain dimensions may be useless for a given dictionary entry and the dimension-wise approach can select only the relevant dimensions.
        Then, the refinement part is trained with sparse vectors which contain only the relevant information.
        Moreover, feature-wise selection with post-attention leads to aggregate sparse vector together, which might bring to sub-optimal results because some dimensions are rarely used.
        
        \setlength{\tabcolsep}{6pt}
        \begin{table*}[t]
            \caption{Impact of the pre-attention or post-attention on the performances for the three proposed attention strategies. Reported Recall@1 (R@1) is on the Cub-200-2011 dataset in percent.}
            \centering
            \begin{tabular}{|c|c|c|c|c|c|}\hline
               & & \multicolumn{2}{c|}{Feature} & \multicolumn{2}{c|}{Dimension} \\\hline
               & Baseline & Pre-att & Post-att & Pre-att & Post-att \\\hline
               R@1 & 52.9 $\pm$ 0.2 & 52.3 $\pm$ 0.4 & \textbf{54.0} $\pm$ 0.4 & \textbf{58.3} $\pm$ 0.2 & \underline{57.8} $\pm$ 0.1 \\\hline
            \end{tabular}
            \label{tab:abla_attention}
        \end{table*}
        
        \setlength{\tabcolsep}{6pt}
        \begin{table}[t]
            \footnotesize
            \caption{Impact of the dictionary size for the feature-wise post-attention mapping and for the dimension-wise pre-attention mapping. Reported results are Recall@1 (R@1) on the Cub-200-2011 dataset in percent.}
            \centering
            \begin{tabular}{|c|c|c|c|c|}\hline
                \multicolumn{5}{|c|}{Feat. + Post-att.} \\\hline
                N & 2 & 4 & 8 & 16 \\\hline
                R@1 & 53.4 $\pm$ 0.3 & 53.3 $\pm$ 0.3 & \textbf{54.0} $\pm$ 0.4 & 48.0 $\pm$ 0.4 \\\hline\hline
                \multicolumn{5}{|c|}{Dim. + Pre-att.} \\\hline
                N & 2 & 4 & 8 & 16 \\\hline
                R@1 & 57.1 $\pm$ 0.3 & 57.4 $\pm$ 0.3 & 58.3 $\pm$ 0.2 & \textbf{58.9} $\pm$ 0.3 \\\hline
            \end{tabular}
            \label{tab:abla_dictionary}
        \end{table}

    \subsection{Comparison to ABE}\label{sec:comp_abe}
        In a second time, we want to evaluate the impact of the structural constraints imposed by the dictionary (\autoref{eq:feat_sel} and \autoref{eq:dim_sel}) by comparing our method to ABE \cite{Kim_2018_ECCV}.
        In ABE, the authors show that a M-head approach already provides strong results on the Cars-196 dataset with 76.1\% (+8.9\%) Recall@1 for $M=8$.
        However, this architecture tends to overfit due to the large number of parameters.
        Then, they propose an enhanced version named ABE that takes advantage of attention maps.
        The divergence loss increases the performance from 69.7\% without the divergence loss to 85.2\% (+15.5\%) in Recall@1 (see Table 2 in \cite{Kim_2018_ECCV}).
        
        Among the drawback of the divergence loss, despite of the additional hyper-parameter, figure that optimizing the loss generates gradients which are in opposition with the metric learning loss ones.
        Indeed, it is designed to reduce the similarity between different branches even when the images are similar.
        We solve this issue in \ourmethod \ where this optimization constraint is replaced by a structural constraint (softmax in \autoref{eq:feat_sel} and \autoref{eq:dim_sel}).
        The orthogonality is ensured by the design of \ourmethod, which allows to simply remove the divergence loss at the price of a reduced expressiveness: feature map entries can be chosen independently in ABE whereas in \ourmethod \ they are constrained to only one dictionary entry.
        In \autoref{tab:CUB-CARS} and \autoref{tab:SOP-INSHOP}, \ourmethod \ performs well compared to ABE with similar results on the Cars-196 dataset (-0.2\% compared to ABE) and higher ones on other datasets such as Cub-200-2011 (+2.2\%), Stanford Online Products (+1.5\%) and Inshop Clothes Retrieval (+3.4\%) for this set of parameters.
    
    \subsection{Dictionary size}\label{sec:abla_dic_size}
        In this section, we evaluate the impact of the dictionary size on \ourmethod.
        We evaluate both the feature-wise and the dimension-wise with dictionary sizes in $\{2, 4, 8, 16\}$.
        Recall@1 on the Cub-200-2011 dataset are reported in \autoref{tab:abla_dictionary}.
        
        In post-attention with feature-wise selection, extreme combinations (\emph{e.g.}, with 16 branches with 32 dimensions) lead to results under the baseline.
        Thus, to increase the performances of such attention strategy, there is a compromise between the representation size of each branch and the number of branches (+$1.1\%$ over the baseline).
        In pre-attention with dimension-wise selection, all parameter combinations for this approach lead to better results than the baseline (+$4.9\%$ to $+5.4\%$).
        The number of branches increases the performance with the log of the dictionary size on the contrary to the previous strategy.

\section{Comparison to the state-of-the-art}\label{sec:results}
    In this section, we compare \ourmethod \ to the state-of-the-art on 4 DML datasets, named Cub-200-2011 \cite{CUB_200_2011}, Cars-196 \cite{CARS_196}, Stanford Online Products \cite{Song_2016_CVPR} and In-Shop Clothes Retrieval \cite{Liu_2016_CVPR_INSHOP}.
    For Cub-200-2011, Cars-196 and Stanford Online Products, we follow the standard splits from \cite{Song_2016_CVPR} and for In-Shop Clothes Retrieval we follow the one from \cite{Liu_2016_CVPR_INSHOP}.
    Especially, the Cub-200-2011 training set is composed of the first 100 classes (\emph{i.e.}, the training and the validation sets from the ablation studies) for a total of 5864 images and its testing set is composed of the last 100 classes for a total of 5924 images.
    The Cars-196 training set is composed of the first 98 classes for a total of 8054 images and its testing set is composed of the last 98 classes for a total of 8131 images.
    The Stanford Online Products training set is composed of 11318 classes for a total of 59551 images and its testing set is composed of 11316 classes for a total of 60502 images.
    Finally, the In-Shop Clothes Retrieval training set is composed of 3997 classes for total of 25882 images and its testing set is composed 3985 classes and it is split into query set of 17218 images and a collection set of 12612 images.
    We report the Recall@K which evaluates, for a given query, if there is at least one image with the same label in the top-K retrieved images.
    We use $K \in \{ 1, 2, 4, 8, 16, 32 \}$ for Cub-200-2011 and Cars-196, $K \in \{ 1, 10, 100, 1000 \}$ for Stanford Online Products and $K \in \{ 1, 10, 20, 30, 40, 50 \}$ for In-Shop Clothes Retrieval.
    
    We report the results for Cub-200-2011 and Cars-196 in \autoref{tab:CUB-CARS} and in \autoref{tab:SOP-INSHOP} for Stanford Online Products and In-Shop Clothes Retrieval.
    \ourmethod \ consistently improves the already strong baseline on the four datasets and for three different loss functions.
    \emph{E.g.}, using the binomial loss, the baseline is improved from 59.6\% to 63.9\% (+4.3\%) on Cub-200-2011 and from 78.8\% to 85.4\% (+6.6\%) on the Cars-196 dataset.
    The same observation is made for both the other loss functions on these datasets.
    
    Moreover, \ourmethod \ leads to better results when compared to the similar approach ABE.
    Their best reported approach, ABE-8, is outperformed by \ourmethod \ with $N=8$ and $c=64$ (total dimension 512) by 2.2\% in R@1 on Cub-200-2011, by 1.5\% on Stanford Online Products and by 4\% on In-Shop Clothes Retrieval.
    We also report results with $N=16$ and $c=32$ (total dimension 512) which are further improved on these datasets: 1.1\% in Recall@1 on the Cub-200-2011 and by 0.4\% on the Cars-196 dataset, leading to the state-of-the-art on the four deep metric learning datasets.

\section{Conclusion}\label{sec:ccl}
    In this paper, we have presented a dictionary-based attention method named \ourmethod \ which consistently improves DML models.
    An ablation study is undertaken to evaluate the benefits of the feature-wise and the dimension-wise selections for two attention strategies named pre-attention and post-attention.
    We show that \ourmethod \ consistently outperforms the baseline for three different loss functions on four datasets (Cub-200-2011, Cars-196, Stanford Online Products, and Inshop Clothes Retrieval).
    Moreover, it outperforms the current state-of-the-art methods on the four datasets.

\section*{Acknowledgments}
    Authors would like to acknowledge the COMUE Paris Seine University, the Cergy-Pontoise University and M2M Factory for their financial and technical support.

\bibliographystyle{apalike}
\bibliography{biblio}

\end{document}